\documentclass[letterpaper, 10 pt, conference]{ieeeconf}  

\IEEEoverridecommandlockouts                              
\usepackage[utf8]{inputenc}
\usepackage[T1]{fontenc}

\usepackage{amsmath}
\usepackage{amsfonts}
\usepackage{amssymb}

\usepackage{amsthm}
\usepackage[ruled,vlined]{algorithm2e}
\usepackage{mathtools}
\usepackage{tabularx}
\usepackage{graphicx}
\usepackage{subfigure}
\usepackage{enumerate}
\usepackage{float}
\usepackage{url}
\usepackage{verbatim}
\usepackage{stfloats}
\usepackage{bbding}
\usepackage{multirow}

\usepackage[normalem]{ulem}
\usepackage{times}
\usepackage{xcolor}
\usepackage{cancel}

\newcommand{\delete}[1]{{\bgroup\markoverwith{\textcolor{red}{\rule[0.5ex]{2pt}{0.4pt}}}\ULon{#1}}}
\newcommand{\deletefig}[1]{{\bgroup\markoverwith{\textcolor{red}{\rule[2.5ex]{2pt}{2.0pt}}}\ULon{#1}}}
\definecolor{rblue}{rgb}{0,0.5,1}

\makeatletter
\let\NAT@parse\undefined
\makeatother
\usepackage[colorlinks,linkcolor=red,anchorcolor=blue,urlcolor=magenta,citecolor=blue]{hyperref}

\usepackage{booktabs}
\usepackage{threeparttable}  
\usepackage{multirow}
\usepackage{pifont}  
\usepackage{algorithm2e}
\usepackage{graphicx}

\usepackage{xcolor}
\definecolor{revised_color}{HTML}{00008B}

\usepackage{cite}

\graphicspath{{pics/}}

\title{\LARGE \bf
SF-TIM: A Simple Framework for Enhancing Quadrupedal Robot Jumping Agility by Combining Terrain Imagination and Measurement
}
\author{Ze Wang$^{1,4}$, Yang Li$^{4}$, Long Xu$^{3}$, Hao Shi$^{1}$, Zunwang Ma$^{4}$, Zhen Chu$^{4}$, Chao Li$^{4}$,\\Fei Gao$^{3}$, Kailun Yang$^{2}$, and Kaiwei Wang$^{1}$%
\thanks{This was supported in part by Zhejiang Provincial Natural Science Foundation of China (Grant No. LZ24F050003), in part by the National Natural Science Foundation of China (Grant No. 12174341 and No. 62473139), in part by the National Key RD Program (Grant 2022YFB4701400), in part by the ``Leading Goose'' R\&D Program of Zhejiang (Grant No.2023C01177), in part by the Hunan Provincial Research and Development Project (Grant No. 2025QK3019), and in part by the Open Research Project of the State Key Laboratory of Industrial Control Technology, China (Grant No. ICT2025B20). 
{\textit{(Corresponding authors: Kaiwei Wang and Kailun Yang.)}}}
\thanks{{$^{1}$State Key Laboratory of Modern Optical Instrumentation, Zhejiang University, China}}
\thanks{{$^{2}$School of Artificial Intelligence and Robotics and National Engineering Research Center of Robot Visual Perception and Control Technology, Hunan University, China}}
\thanks{{$^{3}$State Key Laboratory of Industrial Control Technology, Zhejiang University, China}}
\thanks{{$^{4}$DeepRobotics Co. Ltd., China}}
\thanks{Email: wangkaiwei@zju.edu.cn, kailun.yang@hnu.edu.cn.}
}

\begin{document}

\maketitle
\thispagestyle{empty}
\pagestyle{empty}

\begin{abstract}
Dynamic jumping on high platforms and over gaps differentiates legged robots from wheeled counterparts. Dynamic locomotion on abrupt surfaces, as opposed to walking on rough terrains, demands the integration of proprioceptive and exteroceptive perception to enable explosive movements. In this paper, we propose SF-TIM (Simple Framework combining Terrain Imagination and Measurement), a single-policy method that enhances quadrupedal robot jumping agility, while preserving their fundamental blind walking capabilities. In addition, we introduce a terrain-guided reward design specifically to assist quadrupedal robots in high jumping, improving their performance in this task. To narrow the simulation-to-reality gap in quadrupedal robot learning, we introduce a stable and high-speed elevation map generation framework, enabling zero-shot simulation-to-reality transfer of locomotion ability. Our algorithm has been deployed and validated on both the small-/large-size quadrupedal robots, demonstrating its effectiveness in real-world applications: the robot has successfully traversed various high platforms and gaps, showing the robustness of our proposed approach. A demo video has been made available at~\url{https://flysoaryun.github.io/SF-TIM}.
\end{abstract}

\section{Introduction}
\begin{figure}[t]
	\centering
	\includegraphics[width=0.8\linewidth]{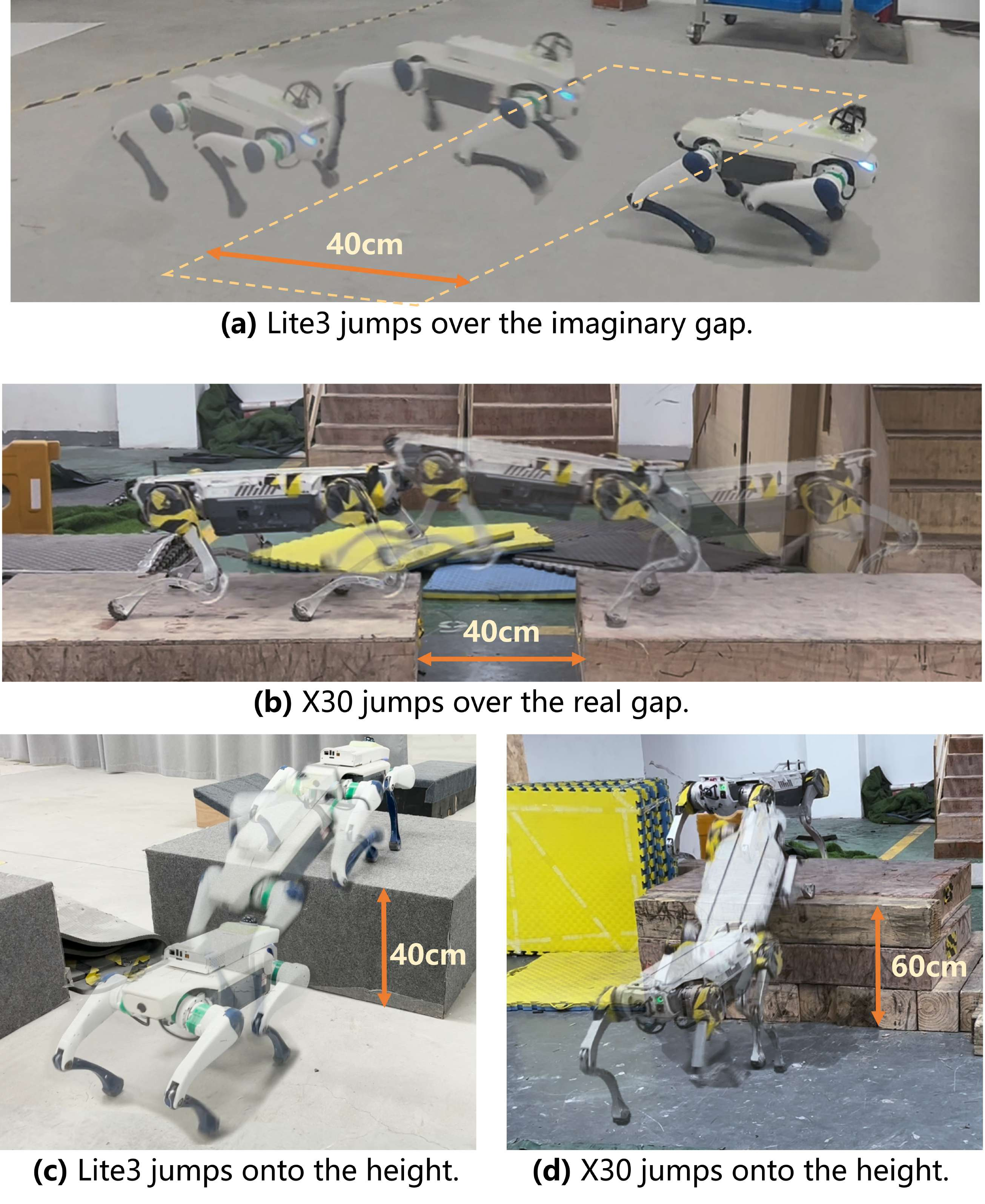}
        \vskip-2ex
         \caption{Jumping experiment of Lite3 and X30 robots. The Lite3 and X30 robots perform horizontal and vertical jumps respectively, with the Lite3 robot jumping over an imaginary gap and the X30 over a real gap.}
	\label{fig:parkour}
\vskip-4ex
\end{figure}

With the rapid development of legged robotics, quadrupedal robots have become essential in exploration and search and rescue missions due to their superior terrain passability~\cite{hoeller2024anymal,nahrendra2023dreamwaq,he2024agile,long2024hybrid,ren2024top,miki2022learning}. 
Unlike wheeled robots, quadrupedal robots excel in handling complex terrains but are relatively difficult to control~\cite{xu2023efficient}.
Reinforcement Learning (RL) algorithms utilizing proprioceptive sensor data, such as Inertial Measurement Units (IMU) and joint encoders, can enhance the terrain adaptability of quadrupedal robots, enabling them to traverse both smooth and rugged terrains and navigate stairs~\cite{nahrendra2023dreamwaq}. However, relying solely on proprioceptive sensors does not enable quadrupedal robots to perform vertical or horizontal jumping maneuvers, which are crucial for enhanced terrain traversal. 
Typically, these robots rely on exteroceptive sensors such as depth cameras or LiDAR~\cite{hoeller2024anymal} to achieve jumping capabilities and further improve their terrain traversal abilities.

Currently, many quadrupedal robot control frameworks rely on exteroceptive sensors like depth cameras and LiDAR.
Depth-camera-based frameworks~\cite{cheng2023extreme,zhuang2023robot} often employ a teacher-student network approach. 
Initially, the teacher network is trained using elevation maps to learn the environmental features. Subsequently, the student network is trained using depth maps under the guidance of the teacher network, transferring learned knowledge through a distillation process.
Due to the substantial memory consumption of depth rendering in Isaac Gym simulator~\cite{makoviychuk2021isaac}, the number of robots trained concurrently is relatively small, leading to higher training costs. 
Additionally, depth cameras usually operate at lower frame rates, necessitating consideration of camera latency, which significantly increases the training overhead. The student model usually doesn't exceed the performance of the teacher model. Although the PIE framework~\cite{luo2024pie} achieves scalable robotic training through NVIDIA Warp-accelerated parallel depth mapping, the computational efficiency is still constrained by the neural network-intensive terrain feature extraction from depth data, which can lead to longer training durations.
The LiDAR-based approach presented by Hoeller~\textit{et al.}~\cite{hoeller2024anymal} is also highly effective and excels in parkour tasks, demonstrating the capability to handle diverse terrains with great proficiency. Due to its goal of enabling robust navigation, this framework is relatively complex, involving multiple modules for perception, navigation, and locomotion, each trained independently. The perception module leverages transformer networks for LiDAR point cloud processing, the navigation module formulates strategies based on the perception data, and the locomotion module executes strategies for tasks such as climbing, jumping, and crouching.

To address the aforementioned issues and to enhance the terrain traversal abilities of quadrupedal robots, we propose SF-TIM, a simple control framework for terrain imagination and measurement. 
Compared with depth-camera frameworks, our approach does not require distillation, significantly reducing training time. This reduction is due to our ability to directly utilize elevation maps during real-world deployment.
Additionally, our elevation maps operate at a frequency of $200\text{Hz}$, minimizing errors introduced by latency. 
Our framework enables a single network to achieve various maneuvers, including climbing upwards, jumping downwards, horizontal jumping, ascending and descending stairs, and controlling locomotion on relatively flat terrain. For jumping maneuvers, we align the robot's heading velocity to the terrain's direction of traversal via remote control and manage its forward speed along the $x$ axis to navigate through the terrain. For other types of terrain, remote commands allow for the adjustment of the robot's velocities in the $x$ and $y$ directions, as well as its angular velocity about the $z$-axis.
We also propose a terrain-guided reward approach specifically to enhance the jumping performance of quadrupedal robots, endowing them to achieve higher terrain levels in simulation. 
To reduce the sim-to-real gap, we introduce a stable and high-speed elevation map generation framework, facilitating zero-shot sim-to-real transfer of locomotion ability.

In summary, our contributions are as follows:
\begin{itemize}
\item We propose \emph{SF-TIM}, a robust terrain-guided LiDAR-based framework using terrain imagination and measurement.
\item We introduce a terrain-guided reward approach to enhance the jumping performance of quadrupedal robots and develop a stable and high-speed elevation map generation framework to reduce the sim-to-real gap, enabling zero-shot sim-to-real transfer.
\item Our approach simplifies quadrupedal robot training by using a single network trained solely with elevation maps. This network enables effective traversal of stairs and maintains control over horizontal speed and z-axis angular velocity in non-jumping scenarios.
\end{itemize}

\section{Related Work}
This section provides a concise review of notable works related to proprioceptive and exteroceptive sensors quadrupedal robot control frameworks.

\subsection{Learning Quadrupedal Robot Locomotion Using Proprioceptive Sensors Only}

This subsection discusses approaches where quadrupedal robots rely exclusively on proprioceptive data. As a result, these robots perceive terrain primarily through contact, using leg or body collisions to detect features such as stairs.

Kumar~\textit{et al.}~\cite{kumar2021rma} proposed Rapid Motor Adaptation (RMA), which enables quadruped robots to adapt in real-time to various challenging terrains without prior exposure during training. 
Wu~\textit{et al.}~\cite{wu2023learning} introduced a locomotion system using Adversarial Motion Priors that enables quadruped robots to traverse challenging terrains robustly and rapidly with only proprioceptive sensors. 
Long~\textit{et al.}~\cite{long2024hybrid} introduced the Hybrid Internal Model (HIM), which leverages the robot’s response to disturbances for robust state estimation, enabling efficient learning and agile locomotion across diverse terrains with minimal sensor input. 
Margolis~\textit{et al.}~\cite{margolis2024rapid} presented
an end-to-end learned controller that achieves record agility for
the MIT Mini Cheetah.
{Atanassov~\textit{et al.}~\cite{atanassov2024curriculum} proposed a curriculum-based reinforcement learning method for quadrupedal jumping that enables dynamic jumping without reference trajectories, achieving a $90cm$ forward jump and versatile omnidirectional motions in real-world experiments.}
Nahrendra~\textit{et al.}~\cite{nahrendra2023dreamwaq} proposed DreamWaQ, which uses deep reinforcement learning with implicit terrain imagination to enable quadrupedal robots to traverse challenging terrains with limited sensing modalities. 
Inspired by DreamWaQ, we introduce terrain imagination into our framework to accelerate agent learning. 
To further unlock jumping capabilities, we incorporate direct terrain measurement into the framework, integrating it with imagination. This integration allows the agent to adapt to various terrains more effectively, with the aim of improving the locomotion stability.

\subsection{Learning Quadrupedal Robot Locomotion Using Exteroceptive Sensors}

By incorporating exteroceptive sensors such as depth cameras or LiDAR, quadrupedal robots can perceive terrain not only through direct contact or collision but also through advanced sensing capabilities.

Cheng~\textit{et al.}~\cite{cheng2023extreme} developed an approach for legged robots to perform extreme parkour by initially training a neural network using elevation maps and then employing a teacher-student method for distillation to operate on depth images from a front-facing camera, enabling precise athletic behaviors despite imprecise actuation and sensing. Zhuang~\textit{et al.}~\cite{zhuang2023robot} developed an end-to-end vision-based system for quadrupedal robots to autonomously learn and execute diverse parkour skills by training each skill individually and then fusing them into a single policy, enabling navigation of complex environments without reference motion data. Hoeller~\textit{et al.}~\cite{hoeller2024anymal} developed a fully learned approach for agile navigation in quadrupedal robots, combining a high-level policy that selects and controls locomotion skills with a perception module for reconstructing obstacles from noisy sensory data, enabling the robot to navigate challenging parkour scenarios without expert demonstrations or prior environment knowledge.

Frameworks relying solely on internal proprioceptive sensors have limited capabilities in unlocking the full potential of quadrupedal robot motion. Existing exteroception-based methods also face various challenges, which are relatively difficult, time-consuming, and involve complex systems. 
Depth map approaches~\cite{cheng2023extreme,zhuang2023robot} that train with elevation maps and then switch to depth maps during training are costly and must account for depth map latency, further increasing training costs.  Luo~\textit{et al.}~\cite{luo2024pie} proposed the Parkour with Implicit-Explicit Learning (PIE) framework, a dual-level estimation approach enabling quadruped robots with low-cost depth sensors to achieve robust parkour performance on challenging terrains.
Methods using LiDAR sensors~\cite{hoeller2024anymal} involve training multiple skills and adding a navigation module for policy switching, which heavily relies on the robustness of the navigation module and receives external inputs such as global position and time command, resulting in a relatively complex system. 
To address these issues, we propose a direct training method using elevation maps, which is more cost-effective and enables a single network to train for vertical and horizontal jumping capabilities. 
This approach enables control of the robot’s velocities in the $x$ and $y$ directions and its angular velocity about the $z$ axis, while also preserving the ability to handle stairs, without requiring distillation or policy switching.

\begin{figure*}[t]
	\centering
	\includegraphics[width=0.9\linewidth]{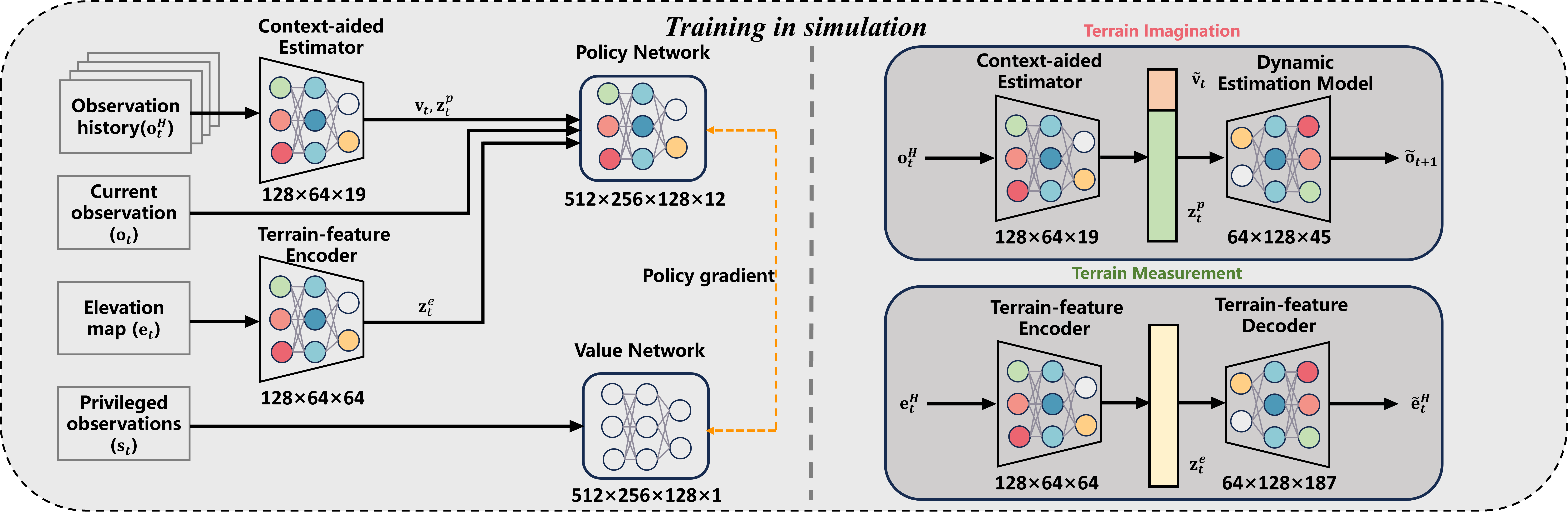}
 \caption{Overview of the training of SF-TIM. The left side illustrates the actor-critic architecture, which includes the policy network responsible for action selection and the value network responsible for evaluating the expected rewards of states. The right side shows the supervision of the network training process for CENet and the Terrain-feature Encoder.
}
	\label{fig:train_flow}
\vskip-2ex
\end{figure*}

\section{SF-TIM: Proposed Framework}
SF-TIM aims to enhance the agility and jumping performance of quadrupedal robots by integrating terrain measurement with imaginative processing within a unified network.

\subsection{Original Inputs of the Agent}

Original inputs of the agent include observation $\mathbf{o}_t$, observation history $\mathbf{o}_t^H$, and elevation map $\mathbf{e}_t$.
The observation vector at time \( t \) is defined as:
\begin{equation}
\mathbf{o}_t = \begin{bmatrix}
\boldsymbol{\omega}_t &
\mathbf{g}_t &
\mathbf{c}_t &
\boldsymbol{\theta}_t &
\dot{\boldsymbol{\theta}}_t &
\mathbf{a}_{t-1}
\end{bmatrix}^T.
\end{equation}
Here, \( \boldsymbol{\omega}_t \) represents the body angular velocity, \( \mathbf{g}_t \) is the gravity vector in the body frame, {\( \mathbf{c}_t \) denotes the remote control's desired command, including \( v_x \), \( v_y \), and \( \omega_z \)}, \( \boldsymbol{\theta}_t \) corresponds to the joint angle, \( \dot{\boldsymbol{\theta}}_t \) is the joint angular velocity, and \( \mathbf{a}_{t-1} \) indicates the previous action. {The observation vector \( \mathbf{o}_t \) is sampled at a frequency of 50Hz.}

{We define a temporal observation vector at time \( t \) as:}
\begin{equation}
{\mathbf{o}_t^H = \begin{bmatrix}
\mathbf{o}_t &
\mathbf{o}_{t-1} &
\dots &
\mathbf{o}_{t-H}
\end{bmatrix}^T.}
\end{equation}
{This vector captures the current and past \( H \) observations, forming a sequence of length \( H+1 \). We set \( H = 5 \) to enhance training efficiency and robustness.}

The elevation map \( \mathbf{e}_t \) is an exteroceptive input representing a scan of the robot’s environment. 

\subsection{Actor and Critic}

To enhance quadrupedal robot jumping agility, we employ an asymmetric actor-critic architecture~\cite{pinto2017asymmetric}, considering that the interplay between the policy and value networks in actor-critic algorithms is sufficient to develop a robust locomotion policy. This architecture is capable of implicitly inferring privileged observations from partial temporal observations and elevation maps, as depicted in Fig.~\ref{fig:train_flow}.

{The policy network, represented as \( \pi_{\phi}(\mathbf{a}_t | \mathbf{o}_t, \tilde{\mathbf{v}}_t, \mathbf{z}^p_t, \mathbf{z}^e_t) \), is a neural network parameterized by \( \phi \). This network determines an action \( \mathbf{a}_t \) based on the proprioceptive observation \( \mathbf{o}_t \), estimated velocity in the body frame \( \tilde{\mathbf{v}}_t \), the latent state of the proprioceptive sensor (\( \mathbf{z}^p_t \)), which implicitly encodes both terrain and robot states, and the latent state of the exteroceptive sensor (\( \mathbf{z}^e_t \)), which implicitly encodes only terrain information. The policy is optimized using the Proximal Policy Optimization (PPO) algorithm~\cite{schulman2017proximal}.}

The action space is represented by a $12$-dimensional vector, $\mathbf{a}_t$, which corresponds to the desired joint angles of the robot. 
To streamline the learning process, the policy is trained to predict the desired joint angles relative to the robot's default standing pose, $\mathbf{\theta}_{\text{stand}}$. 
Therefore, the desired joint angles are given by the following equation:
\begin{equation}
\boldsymbol{\theta}_{\text{des}} = \boldsymbol{\theta}_{\text{stand}} + \boldsymbol{a}_t.
\end{equation}
{Each joint's desired angles are then tracked using a Proportional-Derivative (PD) controller, with the desired joint velocity set to zero.}

The value network is structured to provide an estimation of the state value, $V(\mathbf{s}_t)$. 
In contrast to the policy network, the value network receives a privileged observation, $s_t$, defined as follows:
\begin{equation}
\mathbf{s}_t = \begin{bmatrix} \mathbf{o}_t & \mathbf{v}_t & \mathbf{d}_t & \mathbf{e}_t \end{bmatrix}^T,
\end{equation}
where $\mathbf{d}_t$ is the disturbance force applied arbitrarily to the robot’s body, and $\mathbf{e}_t$ is the elevation map scan of the robot’s environment, acting as an exteroceptive input. 
Within the SF-TIM framework, the policy network is trained to implicitly infer $\mathbf{e}_t$ using proprioceptive data.

\subsection{Combination of Terrain Imagination and Measurement}

{The Context-aided Estimator Network (CENet) is used to transform \(\mathbf{o}_t^H\) into \(\tilde{\textbf{v}}_t\) and the proprioceptive sensor latent state \(\mathbf{z}^p_t\). The terrain-feature encoder transforms \(e_t\) into the terrain-feature vector \(\mathbf{z}_t^e\).}

Inspired by DreamWaQ~\cite{nahrendra2023dreamwaq}, we employ a context vector $ \mathbf{z}^p_t$, which encapsulates a latent representation of the world state. This context vector facilitates the integration of temporal and observational data, enhancing the robustness and adaptability of our approach. However, context vector $\mathbf{z}_t^p$ usually only reflects the terrain information around the robot, especially the area under the feet, but this information usually cannot enhance the robot's jumping agility. Therefore, we introduce terrain-feature vector $\mathbf{z}_t^e$ to allow the policy to stimulate the potential of jumping.

{CENet is capable of estimating both the robot’s forward and backward dynamics as well as a latent representation of the environment. 
It employs a single encoder and a multi-head decoder architecture, as illustrated in the top-right corner of Fig.~\ref{fig:train_flow}. 
The encoder network transforms $\textbf{o}^H_t$ into   $\tilde{{\textbf{v}}}_t$ and latent $\mathbf{z}^p_t$. 
The first decoder head estimates $\tilde{{\textbf{v}}}_t$, while the second head reconstructs $\tilde{\textbf{o}}_{t+1}$.
We utilize a $\beta$-variational auto-encoder ($\beta$-VAE)~\cite{kingma2013auto,higgins2016beta,burgess2018understanding} framework for the auto-encoder setup.}

The optimization of CENet involves a hybrid loss function:
\begin{equation}
\mathcal{L}_\text{CE}=\mathcal{L}_{\text{est}}+\mathcal{L}_\text{VAE},
\label{eqn1
}
\end{equation}
where $\mathcal{L}_{\text{est}}$ and $\mathcal{L}_\text{VAE}$ represent the losses for body velocity estimation and VAE, respectively.
The body velocity estimation loss, $\mathcal{L}_\text{est}$, is defined using Mean Squared Error (MSE):
\begin{equation}
{\mathcal{L}_\text{est}= \text{MSE}(\tilde{\textbf{v}}_t,\textbf{v}_t),}
\label{eqn2
}
\end{equation}
{where $\tilde{\textbf{v}}_t$ is the estimated body velocity and ${\textbf{v}}_t$ is the ground truth from the simulator. The velocity state estimation plays a crucial role in terrain traversal~\cite{ji2022concurrent}.}
The VAE loss, $\mathcal{L}_\text{VAE}$, is formulated as:
\begin{equation}
\mathcal{L}_\text{VAE} = \text{MSE}(\tilde{\textbf{o}}_{t+1},\textbf{o}_{t+1}) + \beta D_\text{KL}(q(\mathbf{z}^p_t|\textbf{o}^H_t) \parallel p(\mathbf{z}^p_t)),
\label{eqn3
}
\end{equation}
where $\tilde{\textbf{o}}_{t+1}$ is the {reconstructed next observation}, $q(\mathbf{z}^p_t|\textbf{o}^H_t)$ is the posterior distribution of $\mathbf{z}^p_t$ given $\textbf{o}^H_t$, and $p(\mathbf{z}^p_t)$ is the prior distribution (a standard normal distribution in this case).
The reconstruction loss is computed using MSE, while the KL divergence serves as the latent loss in the VAE training process. 
This approach ensures effective encoding of $\textbf{o}^H_t$ into meaningful latent representations $\tilde{\textbf{v}}_t$ and $\mathbf{z}^p_t$, thereby enhancing the robustness of CENet for state estimation tasks. Since CENet generally relies on past historical states to obtain the current terrain's implicit representation or to predict future terrain, such as in stair scenarios, it cannot effectively predict jumping scenarios based on past information. 

{To address the aforementioned issue, we incorporate terrain measurement observations and use a terrain-feature encoder to extract terrain features.
The terrain-feature Encoder network, as shown in the bottom-right corner of Fig.~\ref{fig:train_flow}, transforms $\mathbf{e}^H_t$ into terrain-feature latent $\mathbf{z}^p_t$ and the terrain-feature decoder network transforms $\mathbf{z}^p_t$ into $\tilde{\mathbf{e}}^H_t$.}
The elevation map reconstruction loss is also defined using MSE:
\begin{equation}
{\mathcal{L}_\text{terrain}= \text{MSE}(\tilde{\mathbf{e}}^H_t,\mathbf{e}^H_t),}
\label{eqn4
}
\end{equation}
where $\tilde{\mathbf{e}}^H_t$ is the estimated elevation map and ${\mathbf{e}}^H_t$ is the ground truth from the simulator.

\subsection{Reward Function}

Given the goal of enhancing the jumping performance of the quadruped robot, certain reward functions have been refined to address various terrain categories, such as omitting penalties for the robot's y-axis angular velocity and pitch angle. 
It comprises task rewards for tracking the commanded velocity and stability rewards to ensure stable and natural locomotion behavior. 
The specifics of the reward function are detailed in Table~\ref{tab:reward}. 
The total reward for the policy, given an action at each state, is formulated as follows:
\begin{equation}
r_t(s_t, \mathbf{a}_t) = \sum_{i} r_i w_i,
\end{equation}
where $i$ indexes each reward component listed in Table~\ref{tab:reward}, with the rewards for feet edge and feet stumble referring to previous work~\cite{cheng2023extreme}.

\begin{figure}[t!]
	\centering
	\includegraphics[width=0.7\linewidth]{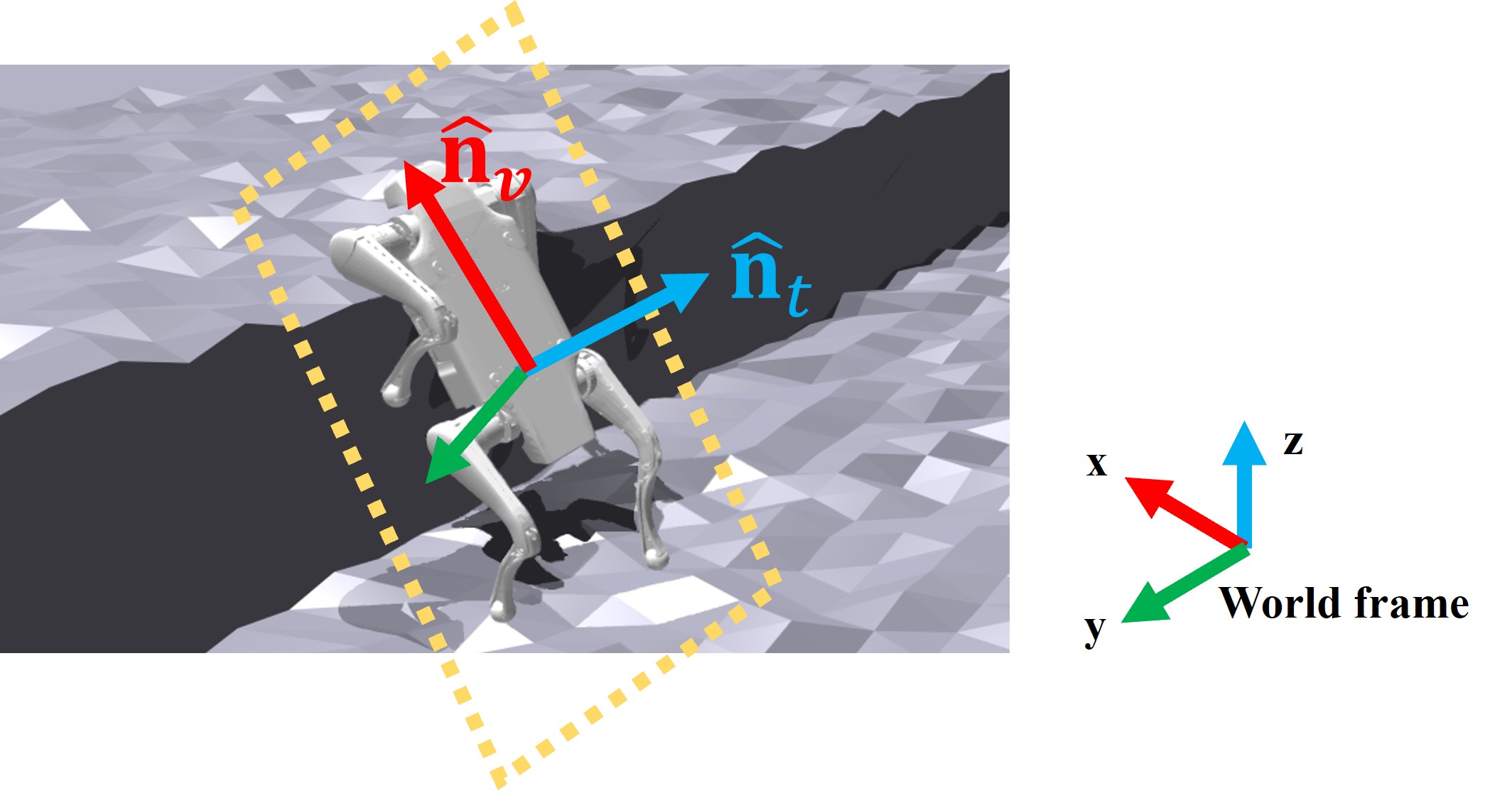}
        \caption{{Terrain-guided reward design. The yellow dotted line represents the plane fitted by the elevation map.}}

	\label{fig:reward}
\end{figure}

{With the quadruped robot's limitations in tracking the x-speed command, it faces challenges when attempting upward jumps on $\tau_5$ terrain. To overcome this, we have designed a reward function that incorporates terrain-specific linear speed tracking. This method uses terrain information to align the robot’s velocity with the direction necessary for effective platform crossing, rather than the default x-direction. We start by selecting the elevation map point set $\mathcal{P}$ within a $1.6m \times 1.0m$ area near the robot's body to fit a plane, defining its normal vector as $\hat{\mathbf{n}}_t$ in Fig.~\ref{fig:reward}. Next, we calculate the direction $\mathbf{\hat{n}}_v$ using terrain orientation:}
\[
R_t = \begin{bmatrix}
\cos(-\arcsin(\hat{\mathbf{n}}_t(0))) & 0 & \sin(-\arcsin(\hat{\mathbf{n}}_t(0))) \\
0 & 1 & 0 \\
-\sin(-\arcsin(\hat{\mathbf{n}}_t(0))) & 0 & \cos(-\arcsin(\hat{\mathbf{n}}_t(0)))
\end{bmatrix},
\]
\[
\mathbf{\hat{n}}_v = R_t \begin{bmatrix}
1\\0\\0
\end{bmatrix}.
\]
{Here, $R_t$ is the rotation matrix with a roll angle ($\Phi$) of $0$, pitch angle ($\Theta$) set to $\arcsin(\hat{\mathbf{n}}_t(0))$, and yaw angle ($\Psi$) of $0$. We then establish a terrain-guided linear velocity tracking (T-L-tracking) reward function, $\min(⟨\mathbf{v}^{world},\mathbf{\hat{n}}_v⟩,{v}^{cmd}_{x})$, which ensures that as the quadruped approaches a platform edge, its velocity direction aligns with $\mathbf{\hat{n}}_v$.}

\begin{table}[t!]
\centering
\footnotesize
\caption{Reward functions and their respective weights.}
\label{tab:reward}
\begin{tabular}{lll}
\toprule
\textbf{Reward} & \textbf{Equation ($r_i$)} & \textbf{Weight ($w_i$)} \\ \hline
T-L-tracking($\tau_5$) & $\min(⟨\mathbf{v}^{world},\mathbf{\hat{n}}_v⟩,{v}^{cmd}_{x})$ & 3.0 \\
L-tracking($\tau_1\sim\tau_4$) & $2 e^{-4(\mathbf{v}^{cmd}_{xy} - \mathbf{v}_{xy})^2}$ & 3.0 \\
A-tracking($\tau_1\sim\tau_5$) & $0.5 \exp(-4(\omega^{cmd}_{yaw} - \omega_{yaw})^2)$ & 0.5 \\
$v_z$($\tau_1\sim\tau_4$) & $-\mathbf{v}_z^2$ & -2.0 \\
$\omega_x$($\tau_1\sim\tau_5$) & $-\omega_{x}^2$ & -0.05 \\
Roll($\tau_1\sim\tau_5$) & $-|\mathbf{g}(0)-{\hat{\mathbf{n}}_t}(0)|^2$ & -10.0 \\
Yaw($\tau_5$) & $-yaw^2$ & -1.0 \\
Joint acc($\tau_1\sim\tau_5$) & $-\dot{\boldsymbol{\theta}}^2$ & $-2.5 \times 10^{-7}$ \\
Body height($\tau_1\sim\tau_5$) & $-(h_{des} - h)^2$ & -10.0 \\
Action rate($\tau_1\sim\tau_5$) & $-(\mathbf{a}_t - \mathbf{a}_{t-1})^2$ & -0.04 \\
Smoothness($\tau_1\sim\tau_5$) & $-(\mathbf{a}_t - 2\mathbf{a}_{t-1} + \mathbf{a}_{t-2})^2$ & -0.03 \\
Hip angle($\tau_1\sim\tau_5$) & $-(\mathbf{d}^{hip}_{des} - \mathbf{d}^{hip})^2$ & -1.0\\
Feet edge($\tau_4$) &  & -10.0\\
Feet edge($\tau_5$) &  & -1.0\\
Feet stumble($\tau_4$) &  &-10.0\\
Feet stumble($\tau_5$) &  &-1.0\\
\bottomrule
\end{tabular}
\end{table}

\begin{figure*}[t!]
	\centering
	\includegraphics[width=0.9\linewidth]{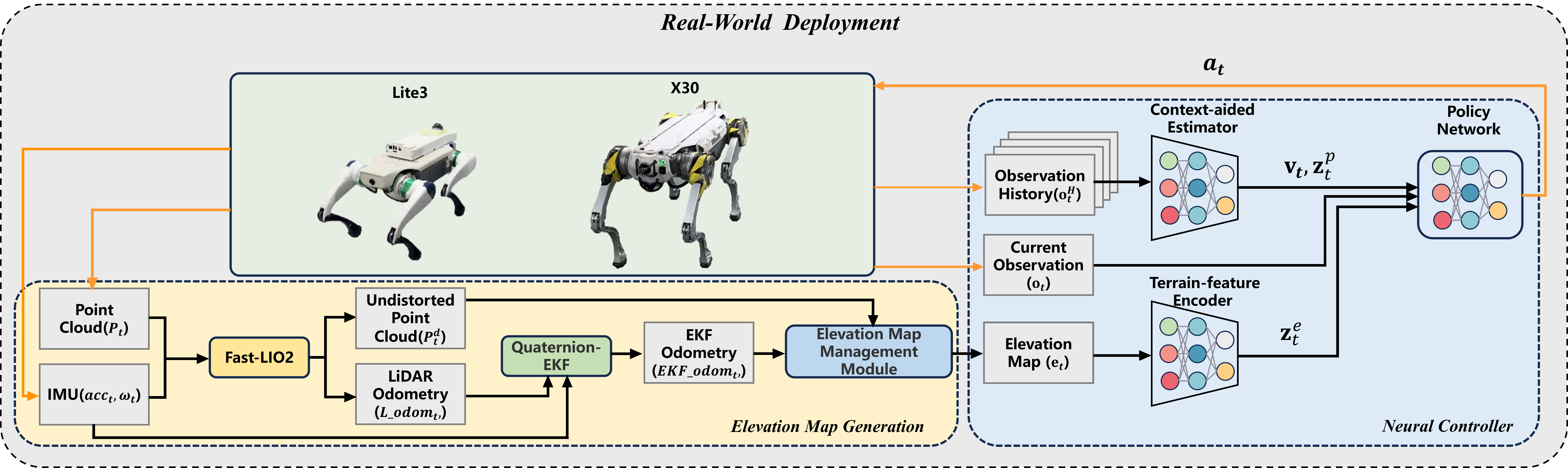}
	\caption{Overview of the real-world deployment of SF-TIM. The yellow section in the lower-left corner represents the elevation map acquisition module $\mathbf{e}_t$, and the right side shows the forward propagation inference of the network.}
	\label{fig:real_flow}
\end{figure*}

\subsection{Curriculum Learning}

\begin{figure}[t!]
	\centering
	\includegraphics[width=0.8\linewidth]{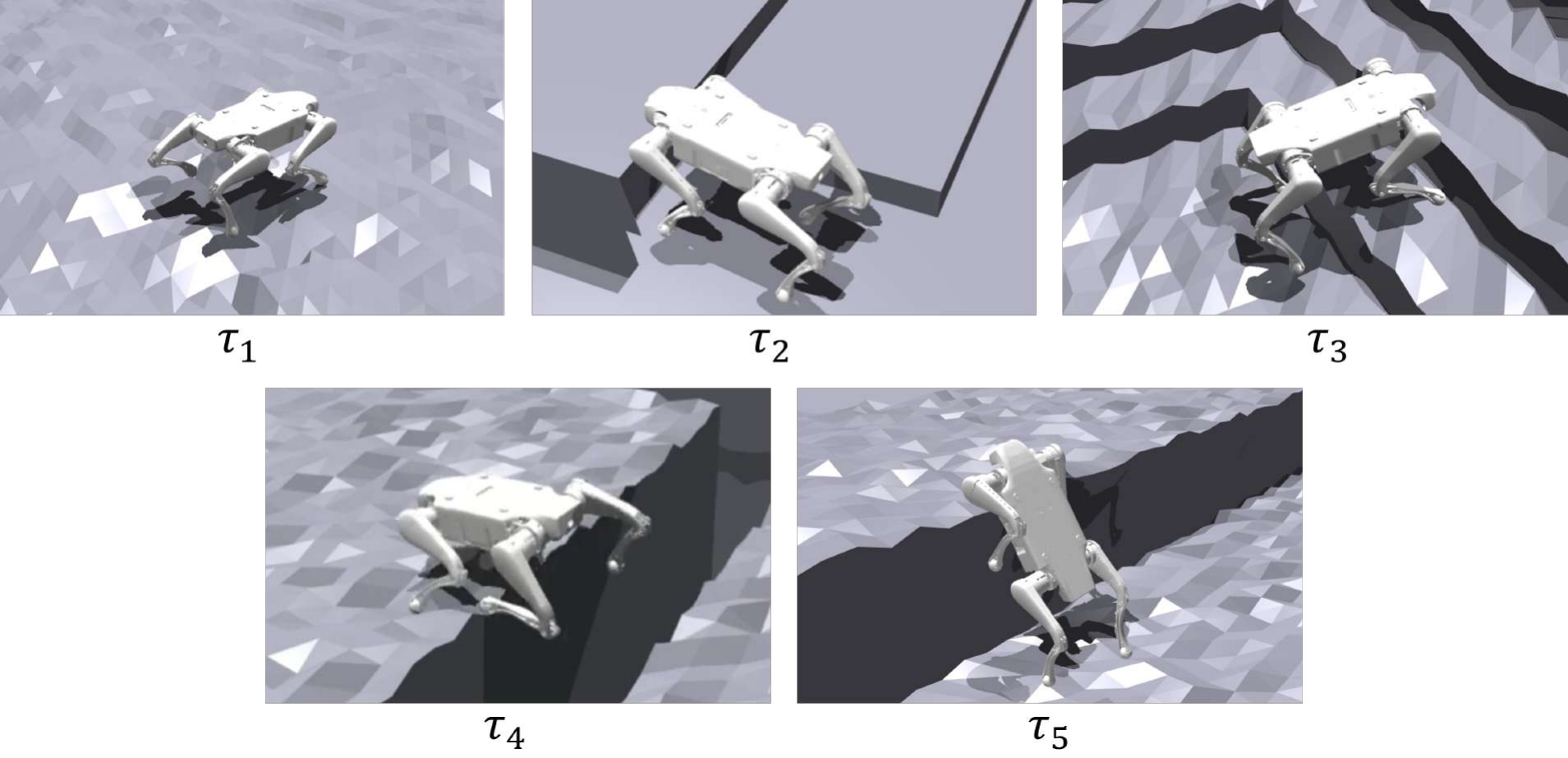}
        \caption{Different terrain types. Slopes: $\tau_1$, Discrete stone: $\tau_2$, Staircases: $\tau_3$, Gaps: $\tau_4$, High platforms: $\tau_5$.}

	\label{fig:terrain}
\end{figure}

We employ simulation-based training methodologies at the Isaac gym facility~\cite{makoviychuk2021isaac}. 
Our approach incorporates a game-inspired curriculum~\cite{rudin2022learning}, which facilitates the incremental acquisition of locomotion policies adept at traversing complex terrains. 
This progressive learning paradigm enhances the robustness and adaptability of the developed locomotion strategies. 
To enable the robot to perform vertical and horizontal movements, as well as navigate stairs and cross flat or small obstacles, we utilize five types of terrain: slopes $\tau_1$, discrete stones $\tau_2$, staircases $\tau_3$, gaps $\tau_4$, and high platforms $\tau_5$, as shown in Fig.~\ref{fig:terrain}. 
The $\tau_4$ and $\tau_5$ require only forward speed commands, whereas the other terrains allow for both forward and lateral speed commands, as well as angular rotation around the z-axis shown in Table~\ref{tab:command}. {When the distance the robot travels forward during the survival time is greater than $0.6$ of the terrain length, it is ensured that the robot has passed the gap and jumped onto the platform, which will increase the difficulty.}

\subsection{Elevation Map Generation}

To minimize the sim-to-real gap during the real-world deployment phase and improve the performance of quadrupedal robots in jumping tasks, we propose a low-latency and high-quality elevation map generation module suitable for SF-TIM, as shown in the lower left corner of the real-world deployment overflow in Fig.~\ref{fig:real_flow}.

We utilize Fast-LIO2~\cite{xu2022fast} to obtain undistorted point clouds and LiDAR odometry at $10\text{Hz}$, using point cloud and IMU data as inputs. For instance, using $10\text{Hz}$ LiDAR odometry at a speed of $1\text{m/s}$ can result in an error of approximately $10\text{cm}$, which can affect the timing of jumps. To obtain higher frequency odometry, we employ a quaternion-based Extended Kalman Filter (EKF) to fuse the LiDAR odometry with IMU data, resulting in $200\text{Hz}$ odometry output. The distortion-corrected point clouds are then fed into the elevation map management module, which maintains global elevation map information. Simultaneously, this module utilizes high-frequency odometry data to generate elevation maps of the quadruped robot's surroundings. Consequently, the acquisition of high-frequency elevation maps reduces the sim-to-real gap.

{The editability of the elevation map module can bring more flexibility. 
For example, we can generate virtual deep trenches in the world coordinate system, so that the robot can jump even on flat ground as shown in Fig.~\ref{fig:parkour}\textbf{(a)}. 
This method not only provides flexibility during debugging but also allows the robot to jump according to user needs, such as visual semantic detection, which enables the robot to jump actively to avoid puddles or more dangerous scenes that cannot be detected by using depth maps. While depth camera images can be considered equivalent to the undistorted point clouds in our system, it is important to note that the accuracy of depth maps is generally not as high as that of LiDAR.
}

\begin{table}[t!]
\centering
\caption{Different command range in different terrain.}
\label{tab:command}
\begin{tabular}{lll}
\toprule
\textbf{Command} & \textbf{Terrain type} & \textbf{Range} \\ \hline

Linear velocity x & $\tau_1\sim\tau_3$ & $[-1.2,1.2]$ \\
Linear velocity x & $\tau_4\sim\tau_5$ & $[0.3,1.2]$ \\
Linear velocity y & $\tau_1\sim\tau_3$ & $[-1.2,1.2]$ \\
Linear velocity y & $\tau_4\sim\tau_5$ & $0$ \\
Angular velocity z  & $\tau_1\sim\tau_3$ & $[-2.0,2.0]$ \\
Angular velocity z  & $\tau_4\sim\tau_5$ & $0$ \\

\bottomrule
\end{tabular}
\end{table}

\section{Experiments}
We utilize the Isaac Gym simulator, built upon the open-source framework outlined in~\cite{rudin2022learning}, to concurrently train the policy, value, CENet networks~\cite{nahrendra2023dreamwaq} and terrain-feature encoder and decoder networks. The training is conducted in parallel with $4,096$ agents subjected to domain randomization.
Domain randomization is employed to enhance the robustness and generalization of the learned policies by varying environmental parameters during training. 
Table~\ref{table:domain_randomization} details the randomized parameters used. 
All algorithms employ PPO~\cite{schulman2017proximal} for training the policy network, with a clipping range of $0.2$, a generalized advantage estimation factor of $0.95$, and a discount factor of $0.99$. 
The networks are optimized using the Adam optimizer~\cite{diederik2014adam} with a learning rate set to $10^{-3}$.
All training is performed on a desktop PC with an Intel Core i7-14700 CPU @ 3.40 GHz, 32 GB RAM, and an NVIDIA RTX 4090Ti GPU.

Given the complexity of our task, we approach training in two distinct stages. Initially, we focus on teaching the robot to walk on terrains $\tau_1$, $\tau_2$, and $\tau_3$, refining its walking policy ($\mathcal{P}_{trot}$) until it reaches a stable state. 
Following this foundational training, we transition to a more challenging all-terrain ($\tau_1\sim\tau_5$) regimen. 
This two-step approach mitigates risks of the robot attempting to jump prematurely on standard terrains ($\tau_1\sim\tau_3$), such as using a pronking gait, which could occur if proceeding with direct, single-phase training.

\subsection{Qualitative Comparisons with Other Quadrupedal Robot Learning Algorithms}

We quantitatively compare our quadrupedal robot learning algorithms SF-TIM with other known algorithms, as shown in Table~\ref{table:benchmark}. In addition to the quantitative comparison, we now offer a qualitative analysis of SF-TIM in relation to other quadrupedal robot learning algorithms.
{One of the key distinctions in training complexity lies in whether parkour skills are trained separately or as part of a unified policy. As indicated in the table, the majority of approaches, including ours, use a single policy. This allows for greater simplicity during deployment, as there is no need for a policy-switching mechanism, which can introduce challenges such as increased computational overhead and the risk of suboptimal transitions between different skills.
Another critical factor is the control of physical system dynamics, specifically the lateral velocity ($v_y$) and the angular velocity around the z-axis ($\omega_z$). Algorithms that enable control of both velocities, such as SF-TIM and Hoeller~\textit{et al.}, offer more precise maneuverability. This capability is particularly advantageous for tasks requiring continuous adaptation to changing terrain or repetitive actions, like jumping onto a platform. In contrast, approaches that lack these controls are inherently limited in their versatility and responsiveness.
SF-TIM demonstrates a significant advantage in training with fewer computational resources. While our method, along with Hoeller~\textit{et al.}\cite{hoeller2024anymal} and Luo~\textit{et al.}~\cite{luo2024pie}, trains 4096 agents, it requires less GPU memory due to the absence of a point cloud completion network and the direct use of elevation maps, which avoids the need for intensive implicit terrain feature identification.  Our method requires shorter training time, enabling rapid parameter adjustments with minimal cost, thus being ideal for resource-limited environments.}

\begin{table*}[t]

\centering
\caption{{Comparison between SF-TIM and other quadrupedal robot learning algorithms.}}
\vskip -2ex
\label{table:benchmark}
\begin{center}
\scriptsize
\begin{tabular}{llccccccc}
\toprule
Method                 &Exteroceptive sensor & Use one policy  & Control $v_y$ & Control $\omega_z$ &{Number of agents}&  {GPU memory} & {Each iteration time}\\
\toprule
Cheng~\textit{et al.}~\cite{cheng2023extreme}       &  Depth camera                               &    Yes                 &        $\times$                  &                       $\times$   & {192} & {14.3G} & {5.7s}                   \\
Zhuang~\textit{et al.}\cite{zhuang2023robot} &   Depth camera                                     &     Yes                 &         $\times$                    &   $\times$   & {256}            & {19.0G} & {2.1s}                            \\
Luo~\textit{et al.}\cite{luo2024pie} &    Depth camera                      &       Yes              &             $\checkmark$    &  $\checkmark$ & {4096} & {>20G} & {7.2s}\\

Hoeller~\textit{et al.} \cite{hoeller2024anymal}       &  LiDAR                                       &       No               &             \checkmark              &     \checkmark   &{4096}        &{>45G}  &               {$/$}               \\
SF-TIM (ours)                   &    LiDAR                         &       Yes              &             $\checkmark$    &  $\checkmark$ & {4096} & {5.7G} & {1.1s}\\
\bottomrule
\end{tabular}
\end{center}
\vskip -2ex
\end{table*}

\subsection{Terrain-guided Reward Simulation Experiment}
We set up $10$ levels of all terrains ($l\in[0,9]$). 
The level table of different terrain parameters is shown in Table~\ref{table:terrain}. 
Among all terrains, $\tau_5$ is relatively more difficult, so we design a terrain-guided tracking velocity reward function specifically for jumping in $\tau_5$. 
For a comparative evaluation, we compare the method with or without our designed terrain guidance reward in the second step of training. 
In the method without our designed reward, the velocity-tracking reward in $\tau_5$ remains the same as that in $\tau_1\sim \tau_4$. 
Due to differences in reward function design, we do not use the reward magnitude for comparison. 
Instead, we compare the average level of the overall terrain after $1000$ iterations, where both networks have converged. 
Additionally, in $\tau_5$, we compare the Success Rates (SR) of different Level 6 (L6) and Level 9 (L9) terrains. 
{The Table~\ref{table:reward_exp} demonstrates that our terrain-guided reward function significantly improves success rates on challenging terrains, with Lite3’s L9 success rate increasing from $15\%$ to $95\%$ and X30’s from $80\%$ to $95\%$. Estimating \(\tilde{\textbf{o}}_{t+1}\) and \(\tilde{\textbf{h}}_{t}\) leads to varying degrees of improvement in the jumping performance of the robots. SF-TIM also outperforms the Extreme Parkour~\cite{cheng2023extreme} teacher model (E-Parkour(T)), especially on difficult terrains, showcasing the advantages of combining TL-tracking and imagination.} {Table~\ref{table:elevation map} additionally compares frequency and compensation methods, with SF-TIM’s 200Hz and historical point cloud compensation offering greater terrain awareness than Hoeller~\textit{et al.}'s 30Hz network-based approach.}

\begin{table}[t!]
\centering
\caption{Terrain parameters of Lite3 and X30.}
\vskip -2ex
\label{table:terrain}
\begin{center}
\scriptsize
\begin{tabular}{llll}
\toprule
Robot    & Terrain    & Terrain parameter & m \\
\toprule
\multirow{5}{*}{Lite3}  & $\tau_1$  & Slope height difference & $0.05\times l$\\
& $\tau_2$  & Discrete stone height & $0.05+0.025\times l$\\
                 & $\tau_3$  & Stair height & $0.05+0.013\times l$\\
                 & $\tau_4$  & Gap width & $0.2+0.035\times l$\\
                  &$\tau_5$  & Platform height & $0.1+0.05\times l$\\
\toprule
\multirow{5}{*}{X30}
 & $\tau_1$  & Slope height difference & $0.05\times l$\\& $\tau_2$  & Discrete stone height & $0.05+0.035\times l$\\
                 & $\tau_3$  & Stair height & $0.05+0.018\times l$\\
                 & $\tau_4$  & Gap width & $0.2+0.06\times l$\\
                  &$\tau_5$  & Platform height & $0.1+0.07\times l$\\
\bottomrule
\end{tabular}
\end{center}
\end{table}

\begin{table}[h]
\centering
\caption{{Performance comparison of SF-TIM ablation experiments and Extreme Parkour.}}
\label{table:reward_exp}
\begin{center}
\setlength{\tabcolsep}{3.5pt}
\scriptsize
\begin{tabular}{llcccc}
\toprule
Robot & Algorithm       & Terrain level & L6 SR(\%) & L9 SR(\%)\\
\toprule

\multirow{5}{*}{Lite3} & SF-TIM    & \textbf{6.0} & \textbf{98} & \textbf{95} \\
                        & SF-TIM  w/o TL-tracking           & 4.8          & 96   & 15  \\  & {SF-TIM w/o $\tilde{\textbf{o}}_{t+1}$}
                           & {5.7}    & {96}    & {85}                   \\
                           & {SF-TIM w/o $\tilde{\textbf{h}}_{t}$}
                           & {5.5}    & {95}    & {82}                   \\
                       \cmidrule{2-5}
                       & {E-Parkour(T)\cite{cheng2023extreme}}   & {5.8}    & {94}    & {90}    \\

\toprule
\multirow{5}{*}{X30}   & SF-TIM      & \textbf{6.1} & \textbf{99} & \textbf{95}  \\  & SF-TIM w/o TL-tracking & 5.5          & 98          & 80          \\ & {SF-TIM w/o $\tilde{\textbf{o}}_{t+1}$} & {5.9}    & {97}    & {86}   \\
 & {SF-TIM w/o $\tilde{\textbf{h}}_{t}$}
                           & {5.6}    & {84}    & {81}     \\
                       \cmidrule{2-5}
                       & {E-Parkour(T)\cite{cheng2023extreme}}   & {6.0}    & {95}    & {93}    \\
\bottomrule
\end{tabular}
\end{center}
\end{table}
\vskip -2ex

\begin{figure*}[t]
	\centering
	\includegraphics[width=0.9\linewidth]{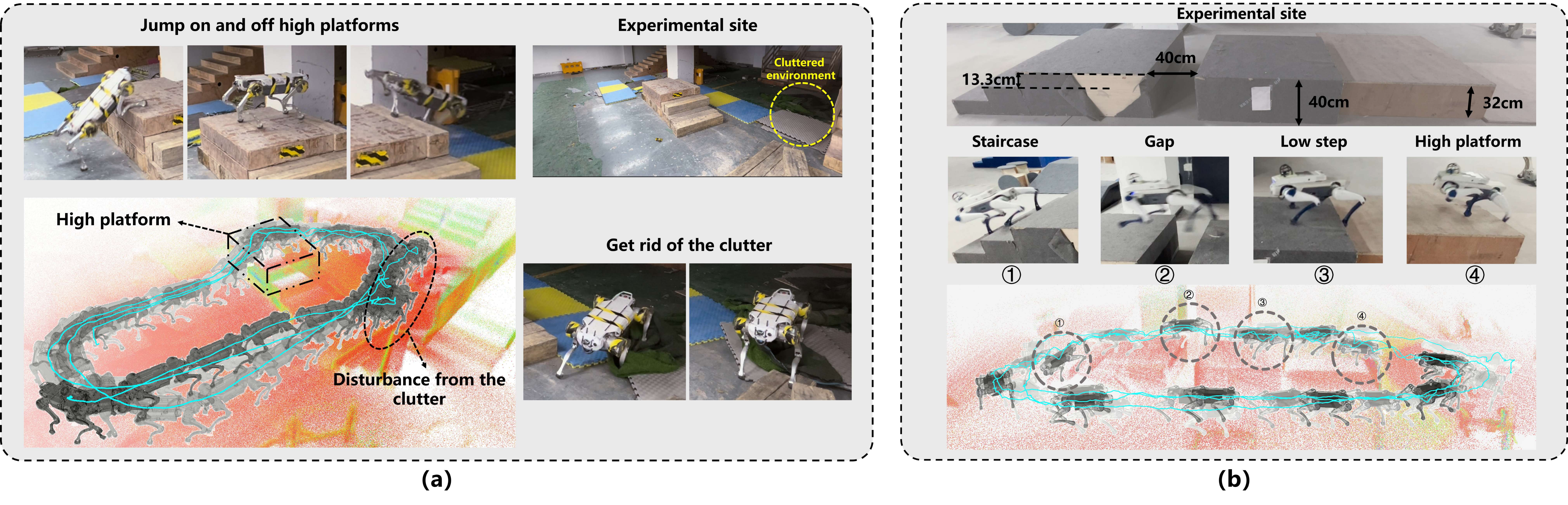}	\caption{\textbf{(a)} Robustness experiment of the high jump platform of the X30 robot. The X30 robot jumps up and down the platform three times consecutively. On the final lap, it entered a cluttered environment where its calf became entangled with flexible debris but successfully got rid of the clutter. \textbf{(b)} Robustness experiment of the various terrains of the Lite3 robot. After circling flat ground once, Lite3 traversed a $32cm$ high platform, an $8cm$ low step, a $40cm$ gap, and a three-step staircase (each step $13.3cm$ high) twice.}
	\label{fig:x30_exp1}
\vskip-2ex
\end{figure*}

\begin{table}[t!]
\footnotesize
\centering
\caption{{Qualitative comparison of elevation map schemes.}}
\vskip -2ex
\label{table:elevation map}
\begin{center}
\setlength{\tabcolsep}{3.4pt}
\scriptsize
\begin{tabular}{lcc}
\toprule
{Method} & {Frequent} & {Compensation method for elevation map}\\
\toprule
{SF-TIM (ours)}               & {200Hz}    & {Historical point cloud}                \\
{Hoeller~\textit{et al.} \cite{hoeller2024anymal} }                & {30Hz}     & {Network}      \\
\bottomrule
\end{tabular}
\end{center}
\vskip -2ex
\end{table}

\subsection{Real-World Experimental Setup}
Real-world experiments were conducted using a Deeprobotics Lite3 robot and an X30 robot. The X30 robot is equipped with four Livox Mid360 LiDARs, while the Lite3 robot is equipped with one Livox Mid360 LiDAR.
The Lite3's elevation map generation module and the motion strategy module are run on NVIDIA NX and RK3588 respectively.
The two parts of the X30 robot run on two separate RK3588 boards. 
Communication between the two boards is achieved using User Datagram Protocol (UDP). During inference, the policy operates synchronously with the CENet at $50\text{Hz}$. The PD controller tracks the desired joint angles using proportional and derivative gains, with $Kp{=}28$ and $Kd{=}0.7$, respectively. For the Lite3 robot, the PD controller gains are $Kp{=}30.0$ and $Kd{=}1.0$, whereas for the X30 robot, they are $Kp{=}120.0$ and $Kd{=}3.0$. 

\begin{table}[t!]
\footnotesize
\centering
\caption{Domain randomization ranges applied in the simulation.}
\vskip -2ex
\label{table:domain_randomization}
\begin{center}
\scriptsize
\begin{tabular}{llc}
\toprule
Parameter                        & Randomization range & Unit \\
\toprule
Payload                     & $[-1, 2]$ & $\mathrm{~kg}$     \\ 
$K_p$ factor   & $[0.9, 1.1]$ & $\mathrm{~Nm/rad}$    \\ 
$K_d$ factor        & $[0.9, 1.1]$ & $\mathrm{~Nms/rad}$   \\ 
Motor strength factor        & $[0.9, 1.1]$ & $\mathrm{~Nm}$   \\ 
Center of mass shift        & $[-50,50]$ & $\mathrm{~mm}$    \\
Friction coefficient       & $[0.2, 1.25]$ & -    \\
System delay        & $[0.0, 15.0]$ &  $\mathrm{~ms}$   \\
Noise ratio in elevation map    & $[0.0, 0.1]$ &  -   \\
Magnitude of noise in the elevation map & $[-1.0, 2.0]$ &  m\\
\bottomrule
\end{tabular}
\end{center}
\vskip -2ex
\end{table}

\subsection{Long-Time Jumping Test and Robustness Analysis}
We deploy our algorithm SF-TIM on X30 and Lite3, respectively, conducting repeated experiments in challenging scenarios such as raised platforms and gap crossings to validate the robustness of our algorithm. 
The experimental results of jumping platforms with X30 are shown in Fig.~\ref{fig:x30_exp1} \textbf{(a)}. 
The X30 robot jumps up and down the platform three times consecutively, with the final landing in a cluttered environment where the quadrupedal robot's calf becomes entangled with flexible debris. 
Despite this, the algorithm's strong robustness allowed us to clear the debris using remote control. 
In the experiment, we employ a single policy to control the robot's steering, forward and backward movements, as well as lateral movement. 

To verify the universality and robustness of our algorithm across different quadruped robots, we have deployed and conducted experiments on the Lite3 quadruped robot. Despite being equipped with only one Livox Mid360 LiDAR, resulting in a smaller perception range compared to the X30, the Lite3 still performed robustly.
Additional experimental sites for Lite3 were established, as depicted in Fig.~\ref{fig:x30_exp1} \textbf{(b)}. Initially, experiments are conducted on flat ground with one complete circuit around the site, followed by two full traversals of the terrain. The terrain is set up sequentially with a $32cm$ high platform, an $8cm$ low step, a $40cm$ gap, and a three-step staircase with each step being $13.3cm$ high. The Lite3 robot successfully traversed this terrain twice and exhibited agile maneuverability on flat ground. 
Overall, our algorithm demonstrates strong robustness and performs effectively in traversing diverse terrains.

\section{Conclusions and Limitations}
In this paper, we present a novel and robust terrain-guided LiDAR parkour framework, denoted as SF-TIM, utilizing elevation maps to address the challenges associated with quadrupedal robot terrain traversal. 
Compared to existing depth-camera parkour frameworks, our approach significantly reduces training time by training only the teacher network. 
The frequency of elevation maps is synchronized with the localization frequency, effectively mitigating latency errors. We have successfully demonstrated climbing, jumping, and traversing various terrains, as well as controlling locomotion on flat terrain, all through a single network. 
The proposed terrain-guided reward approach enhances the jumping performance of quadrupedal robots, facilitating higher terrain level achievements in simulation. 
Furthermore, the integration of a stable and high-speed elevation map generation framework aims to bridge the sim-to-real gap.

Our research primarily focuses on jumping maneuvers and does not encompass a wide range of parkour actions, such as traversing narrow gaps or squeezing through low passages. This limitation arises from our use of a 2.5D elevation map, which provides only a single z-value for each x and y position, lacking information on the upper and lower boundaries. Additionally, we have only utilized approximately $80\%$ of the robot's torque capacity. To fully exploit its performance potential, future training will incorporate constraints based on the motor’s external characteristic curves. To support these actions and enhance mobility, we also plan to increase command channels to control the robot's height in the future.

\bibliographystyle{IEEEtran}
\bibliography{bib}

\end{document}